\documentclass[12pt, epsf, rotating, pdftex, lscape, letterpaper]{article}
\usepackage[margin=0.95in]{geometry} 

\usepackage{lineno, amsmath}
\usepackage{booktabs} 
\usepackage{multirow}
\usepackage{soul} 
\usepackage{microtype}
\usepackage{amsmath}
\usepackage[ruled,linesnumbered,lined,boxed,commentsnumbered]{algorithm2e}
\usepackage{cite}
\usepackage{amsmath,amssymb,amsfonts}
\usepackage[figuresright]{rotating}                                                         
\usepackage{soul}                                                             
\usepackage{microtype}                                                     
\usepackage{caption}                                                        
\usepackage{forest}
\usepackage{blkarray}
\usepackage{comment}

\begin{document}

\begin{center}
{\large \bf Stochastic models of Jaya and semi-steady-state Jaya algorithms}

\vspace{0.4cm}
Uday K. Chakraborty\\

Department of Computer Science\\
University of Missouri, St. Louis, MO 63121, USA\\
chakrabortyu@umsl.edu\\
\end{center}


\noindent 

\abstract{ 
We build stochastic models for analyzing Jaya and semi-steady-state Jaya algorithms. 
The analysis shows that for semi-steady-state Jaya (a) the maximum expected value of the number of worst-index updates per generation is a paltry 1.7 regardless of the population size; (b) regardless of the population size, the expectation of the number of best-index updates per generation decreases monotonically with generations; (c) exact upper bounds as well as asymptotics of the expected best-update counts can be obtained for specific distributions; the upper bound is 0.5 for normal and logistic distributions, $\ln 2$ for the uniform distribution, and $e^{-\gamma} \ln 2$ for the exponential distribution, where $\gamma$ is the Euler-Mascheroni constant; the asymptotic is $e^{-\gamma} \ln 2$ for logistic and exponential distributions and $\ln 2$ for the uniform distribution (the asymptotic cannot be obtained analytically for the normal distribution). The models lead to the derivation of computational complexities of Jaya and semi-steady-state Jaya. The theoretical analysis is supported with empirical results on a benchmark suite. The insights provided by our stochastic models should help design new, improved population-based search/optimization heuristics.
}

\section{Introduction}
\label{sec:introduction}
The Jaya algorithm (technically, heuristic or meta-heuristic) \cite{rao2016jaya, rao2019jaya} is one of the newest members of the evolutionary computation family. This algorithm and its variants have been highly successful in global optimization in continuous domains and have seen wide applicability in diverse areas, including engineering 
\cite{venkata2020multi, tripathy2021jaya, yadav2022effective}, manufacturing \cite{agarwal2018new},
 energy \cite{gupta2021efficient}, fuel cells \cite{chakraborty2019proton}, healthcare \cite{satapathy2018jaya} and finance \cite{mohapatrajaya}.  
A recent survey can be found in \cite{zitar2021intensive}. Within the genetic and evolutionary computation family, Jaya is unique in its use of a minimum number of parameters, a fact that doubtless contributes to this algorithm's popularity among practitioners.  
The semi-steady-state Jaya (SJaya for short) \cite{chakraborty2020semi} has been shown to outperform the standard Jaya on benchmark problems, with the improvement in performance attributed primarily to the new update strategies that SJaya employs for the best and worst members of the population.

Despite their explosive growth, no theoretical analysis of Jaya or its variants has, to our knowledge, been reported in the literature. Such analysis is fundamental to our understanding of why the method works the way it does and is a necessary prerequisite to designing better, newer methods for tackling hard optimization problems.
This paper provides a rigorous theoretical underpinning of this powerful algorithm, modeling the algorithm as a stochastic process and deriving bounds and asymptotics for important performance metrics.
The model allows us to investigate the costs of the update strategies, revealing several interesting facts about the working of Jaya and SJaya, leading to the derivation of the computational complexities of the algorithms. 

The Jaya pseudocode \cite{chakraborty2019proton} and SJaya pseudocode \cite{chakraborty2020semi} are presented as Algorithms~\ref{algooriginal} and \ref{algonew}, respectively.

\begin{algorithm}
initialize the population\;
\While{a pre-determined stopping condition is not satisfied}{
find the best and the worst individuals in the population, and~initialize {\it bestIndex} to the index of the best individual and {\it worstIndex} to the index of the worst individual\;
set the parameters, independently of one another, to~random values between 0.0 and 1.0\;
\For{each individual in the population starting from the first index}{
 create a new individual using the current individual, the~individual at {\it bestIndex}, the~individual at {\it worstIndex}, and~the random parameters\;

\If{the new individual is at least as good as the current individual}{
replace the current individual with the new individual\;

}
}
}
\caption{Jaya}
\label{algooriginal}
\end{algorithm}

\begin{algorithm}
initialize the population\;
find the best and the worst individuals in the population, and~initialize {\it bestIndex} to the index of the best individual and {\it worstIndex} to the index of the worst individual\;
\While{a pre-determined stopping condition is not satisfied}{
set the parameters, independently of one another, to~random values between 0.0 and 1.0\;
\For{each individual in the population starting from the first index}{
 create a new individual using the current individual, the~individual at {\it bestIndex}, the~individual at {\it worstIndex}, and~the random parameters\;

\If{the new individual is at least as good as the current individual}{
replace the current individual with the new individual\;
\If{the current individual is better than the individual at bestIndex}{
update {\it bestIndex} to set it to the current index\;
}
\If{the current individual's index is the same as {\it worstIndex}}{
find the worst individual in the population and set {\it worstIndex} to the index of the worst individual\;  
}
}
}
}
\caption{Semi-steady-state Jaya}
\label{algonew}
\end{algorithm}

\section{Framework for the analysis}
We assume, without loss of generality, an indexed representation (e.g., an array) (Fig.~\ref{fig:indexed}) of the members of the population. The best and the worst members (individuals) are determined with respect to the fitness / utility / cost  or some objective function. A single run of  Jaya or SJaya comprises a number ($G$, say) of generations, and each generation consists of $n$ steps or iterations, where $n$ is the population size.

\begin{figure}
\centering
\begin{tabular}{|c|}    \hline                                      
$n$ \\ \hline
\vdots \\ \hline
$k + 1$ \\ \hline
$k$ \\ \hline
$k -1$ \\ \hline
\vdots\\ \hline
1 \\
\hline
\end{tabular}
\caption{Indexed representation of population members (population size = $n$)}
\label{fig:indexed}
\end{figure}

A single iteration involves determining whether or not the member at index $i$ ($i = 1, 2, \cdots, n$) is to be replaced with a new individual. Clearly, it does not matter whether we traverse the population (array) in a top-to-bottom or bottom-to-top or any other fashion, as long as no index is left out or considered more than once. Suppose, for ease of discussion, we traverse the population in Fig.~\ref{fig:indexed} sequentially from the top (index $n$) to the bottom (index $1$).

\section{Updating SJaya's worst-of-population index}  \label{sec:worstupdate}
Because the population changes with time, the index of the population's worst individual is time-dependent; that is, the worst individual's index may change after every replacement of the current (most recent) worst individual. Thus it is possible for the (current) worst individual to be encountered more than once during the top-to-bottom scan in a given generation of the population. The present analysis assumes that when the worst individual is encountered, it is replaced with a new (better or identical-fitness) individual with probability $p$. We also assume that the value of $p$ does not change during a run.

Let us use the name findWorst() to indicate the function called to find the index of the worst individual in the population, and let worstIndex represent the index of the worst individual at any point in the course of a run. Because a simple linear scan of the population is enough to find the worst fitness, the worst-case complexity of findWorst() is $\Theta(n)$.

Let $X$ be the (discrete) random variable representing the total number of calls, in an entire generation, to the function findWorst(). We are interested in finding the expected value of $X$, because the higher this expectation, the higher the cost of SJaya.

Suppose that at the beginning of a new generation, the worst individual in the entire population is at index $k$, i.e., worstIndex is $k$, with $1 \le k \le n$. During the course of the generation, when this individual at index $k$ is encountered, it will either stay unaltered or be replaced with a new individual. As mentioned earlier, the probability of replacement is assumed to be independent of $k$ and equal to $p$; thus the individual at index $k$ stays unaltered with probability $1-p$.
If it stays unchanged, worstIndex stays unaltered. If, however, it undergoes replacement, we must find (by using a call to findWorst()) which individual in the population is the new worst (it is possible that the newly arrived individual at index $k$, while better than the individual just replaced, turns out to be the worst in the population at that point in time). 
The new worst individual, as identified by the above-mentioned call to findWorst(), must be
\begin{itemize}
\item either in the already-traversed portion of the population array (at an index between $k$ and $n$, inclusive, in Fig.~\ref{fig:indexed});
\item or in the yet-to-be-traversed part of the population (at an index $h$, with $1 \le h \le k-1$).
\end{itemize}
In the first case above, no further call to findWorst() is needed for the rest of the generation, while in the second, the story will repeat itself with the new worst individual, necessitating a total of up to $h$ (i.e., at least zero but at most $h$) further calls to findWorst() for the rest of the current generation. 

Let $W$ be the discrete random variable representing the index of the worst individual in the population at a particular iteration of a particular generation during the execution of a run. Let us use $t$ to represent the iteration number (not to be confused with the generation number for which we will use the notation $g$). Thus $1 \le t \le n$ and $1 \le g \le G$. For  the $n$ steps (iterations) in any generation of the SJaya, the corresponding variables are $W^{(0)}$ (the index of the worst individual at the start of a new generation), $W^{(1)}$ (the worst individual's index after the first iteration of the generation is over), and so on. Thus $W^{(n)}$ of a given generation is the same as $W^{(0)}$ of the immediately following generation. At the very beginning,  under the assumption that the initial population is randomly generated, all slots of the array in Fig.~\ref{fig:indexed} are equally likely to hold the worst one ($P$ stands for probability): 

\begin{equation}
P(W^{(0)} = k|n) = \frac{1}{n} \quad \text{for $k = 1, 2, \cdots, n$}. 
\label{eq:initdistri}
\end{equation}
As the iterations (and generations) roll on, the distribution of the worst individual in the population may deviate from the uniform, depending on the policy used to update the population. That is, for $t > 0$,
we do not in general have a strong reason to assume a uniform distribution for $P(W^{(t)} = k|n)$. Now, for the present analysis, we do not need $P(W^{(t)} = k|n)$ as much as we need the conditional probability  $P(W^{(t+1)} = j | W^{(t)} = k; n)$, which, in the absence of any further information, is assumed to be uniform (at all generations):  
 
\begin{equation}
P(W^{(t+1)} = j | W^{(t)} = k; n) = \frac{1}{n} \quad \text{for any $(j, k)$ pair and any $t$}. 
\label{eq:transition}
\end{equation}

In the course of a generation, when the worst individual is encountered, a new individual is produced and is compared against the worst individual (what happens to the worst individual is no different from what happens to every other individual in the population at the given generation). Now, if the new individual has a fitness that is better than or equal to that of the worst individual, the former replaces the latter, thereby necessitating the finding of which individual in the post-replacement state is the (new) worst in the population. This entails one call to worstFind(). Therefore, in the event of the replacement of the worst individual, at least one call must be made to findWorst(). Thus for an entire generation (recall the definitions of $X$ and $p$), we have

\begin{gather}
P(X > 0| W^{(0)} = k; n)  = p,  \label{eq:x>0}\\
P(X = 0| W^{(0)} = k; n)  =  1 - p. 
\end{gather}

Given $W^{(0)} = k$ for a certain generation (recall that one generation equals $n$ iterations), the variable $X$ can assume one of the following values for that (entire) generation: $0, 1, 2, \cdots, k$. 
Thus equation~\ref{eq:x>0} can be written more specifically as:

\begin{equation}
\sum_{m=1}^{k} P(X = m| W^{(0)} = k; n)  = p.
\end{equation}

Assuming $W^{(0)} = k$, consider the top-to-down journey in Fig.~\ref{fig:indexed}.
The index $k$ may be thought of as indicating the point of demarcation, splitting the population into a top part of size $n-k+1$ and a bottom part of size $k-1$. Given $W^{(0)} = k$, we can describe the result of a call to findWorst() as either an ``up" move (when the index returned by findWorst() is $\ge k$) or a ``down" move (when the returned index is $< k$). Thus, given $W^{(0)} = k$, the event $X=1$ takes place when, starting at index $k$, we either move ``up" once, never to move anywhere else, or move ``down" once and do not move further:
 \begin{eqnarray}
P(X = 1| W^{(0)} = k \ge 1; n)  & = & p \times   \sum_{i=k}^{n}\frac{1}{n} \, + \,  p \times \sum_{i=1}^{k-1}\frac{1}{n} \times (1-p) \nonumber \\
                                              & = & \frac{p}{n} \left(n + p - pk \right).     
\label{eq:m1}
\end{eqnarray}
The event $X=2$ occurs in one of the following two scenarios: (a) starting from slot $k$, the first move is a ``down" move to slot $i \in [1, k-1]$, and the second one is a move ``up" to any slot $\in$ $[i, n]$; and (b) the first two moves are ``down" each, followed by no further movement:
\begin{eqnarray}
P(X = 2| W^{(0)} = k \ge 2; n)  & = & p\sum_{i=1}^{k-1} \left( \frac{1}{n}  \times p \sum_{j=i}^{n}\frac{1}{n} \right)  \, + \,  p\sum_{i=2}^{k-1} \left(\frac{1}{n} \times p \sum_{j=1}^{i-1}\frac{1}{n} \times (1-p) \right)       \nonumber \\
                                    & = & \frac{p^{2}}{n^{2}} (k-1) \left(n +p - \frac{pk}{2}\right).   
\label{eq:m2}
\end{eqnarray}
 
Similarly, $X$ is 3 when we have either (a) a ``down" move from $k$ to any location $i \in [2, k-1]$,  followed by a second  ``down" move from $i$ to any location $j \in [1, i-1]$, followed, finally, by an ``up" move from $j$ to any location $\in [j, n]$; or (b) three successive ``down" moves followed by no further movement:

\begin{align}
P(X = 3| W^{(0)} = k \ge 3; n)   = & p\sum_{i=2}^{k-1} \left( \frac{1}{n}  \times p \sum_{j=1}^{i-1} \left( \frac{1}{n} \times p \sum_{h=j}^{n}\frac{1}{n} \right) \right)  \nonumber \\
                         & + p\sum_{i=3}^{k-1} \left(\frac{1}{n} \times p \sum_{j=2}^{i-1} \left( \frac{1}{n} \times p \sum_{h=1}^{j-1}\frac{1}{n} \times (1-p) \right) \right)       \nonumber \\
                                     = & \frac{p^{3}}{2n^{3}} (k-1)(k-2) \left( n +p  - \frac{pk}{3} \right).   
\label{eq:m3}
\end{align}

\subsection{The general case}
For the general case $X=m$, where $m \in [1,n]$, we have the following theorem (the product notation $\Pi$ evaluates to 1 when the upper bound is less than the lower bound):

\vspace{4mm}
{\bf Theorem 1}: For $m \ge 1$,
\begin{eqnarray}
P(X = m| W^{(0)} = k \ge m; n)  =  \frac{1}{(m-1)!} \; \left(\frac{p}{n}\right)^{m}   
\left(n + p - \frac{pk}{m} \right)   \prod_{j=1}^{m-1} (k-j).  
\label{eq:theorem}
\end{eqnarray}

Proof: We present a proof by induction. The base cases for $m$ = 1, 2 and 3 are already established via equations~\ref{eq:m1}, \ref{eq:m2}, \ref{eq:m3}. 
The proof will be complete when, assuming the theorem is true for $m=q \ge 1$, we show that it is true for $m=q+1$.

Starting from location $k$, any $(q+1)$-move sequence comprises a first ``down" move to any location $i \in [q, k-1]$, followed by a sequence of $q$ further moves, with the first move of the $q$-move sequence starting at location $i$. Thus 

\begin{equation*}
P(X = q+1| W^{(0)} = k \ge q+1; n)   =  p \sum_{i=q}^{k-1} \left( \frac{1}{n} \times P(X = q| W^{(0)} = i \ge q; n)\right).                        
\end{equation*}
Substituting for $P(X = q| W^{(0)} = i \ge q; n)$ from the theorem (equation~\ref{eq:theorem}) into the above equation, we have
\begin{linenomath}\begin{align}
P(X = q+1| W^{(0)} = k \ge q+1; n)  = &  \frac{1}{(q-1)!} \; \left(\frac{p}{n}\right)^{q+1}
\sum_{i=q}^{k-1} \left(     \left(n +p - \frac{pi}{q} \right)
      \prod_{j=1}^{q-1} (i-j)       \right) \nonumber \\
                                    = &  \frac{1}{(q-1)!} \; \left(\frac{p}{n}\right)^{q+1} \Bigg[ (n+p) \sum_{i=q}^{k-1}  \prod_{j=1}^{q-1} (i-j) - \nonumber \\
                                       &  \qquad \qquad \qquad \qquad \frac{p}{q} \sum_{i=q}^{k-1}   \left(i      \prod_{j=1}^{q-1} (i-j) \right)     \Bigg]. 
\label{eq:almostcomplete}
\end{align}\end{linenomath}
Now, it can be shown (after some algebra) that 
\begin{eqnarray*}
\sum_{i=q}^{k-1}  \prod_{j=1}^{q-1} (i-j)   =  \frac{(k-q) \; (k-1)!}{q \; (k-q)!} \quad \text{and}
\end{eqnarray*}
\begin{eqnarray*}
\sum_{i=q}^{k-1}   \left(i  \prod_{j=1}^{q-1} (i-j) \right)   =  \frac{k(k-1)(k-2)\cdots(k-q)}{q+1}.
\end{eqnarray*}
Use of these two identities in equation~\ref{eq:almostcomplete} followed by some simplification yields
\begin{linenomath}\begin{equation*}
P(X = q+1| W^{(0)} = k \ge q+1; n)  =  \frac{1}{q!} \; \left(\frac{p}{n}\right)^{q+1}   \left(n +p - \frac{pk}{q+1} \right)   \prod_{j=1}^{q} (k-j). 
\end{equation*}\end{linenomath}
Q.E.D.

The expectation of $X$, given a particular $W^{(0)}$ and a particular $n$, is now obtained as
\begin{linenomath}\begin{align}
E(X|W^{(0)} = k; n)  = & \; 0 \times (1-p) + \sum_{m=1}^{k} (m \times P(X = m| W^{(0)} = k \ge m; n)) \\
                               = & \sum_{m=1}^{k} (m \times P(X = m| W^{(0)} = k \ge m; n)).
\end{align}\end{linenomath}
Finally, the expectation of $X$, given an $n$, is
\begin{linenomath}\begin{align}
E(X | n) = & \; \sum_{k=1}^{n} E(X|W^{(0)} = k; n) P(W^{(0)} = k | n) \\
            = & \; \frac{1}{n} \sum_{k=1}^{n} E(X|W^{(0)} = k; n).
\label{eq:theorEworst}
\end{align}\end{linenomath}

The probability $P(X = m| W^{(0)} = k \ge m; n)$ and hence the expectation $E(X|n)$ are monotone increasing in $p$, with $E(X|n)$ reaching its 
 highest possible value when $p = 1$. 
Table~\ref{tab:maxE} presents the analytically obtained maximum value of $E(X|n)$ for different values of $n$.  It is interesting to note from Table~\ref{tab:maxE} that the maximum value is almost constant, regardless of the population size. Given the existing body of research on population sizing in evolutionary computation, 
we can say that the spread of population sizes in Table~\ref{tab:maxE} is wide enough to include almost all cases of practical interest. Ignoring less-than-50 values of $n$ as too small, we arrive at the rather remarkable conclusion that the expected number of worst-index updates per generation is 1.7 for almost any population size.

\begin{table}
\caption{Maximum value of $E(X | n)$}
\label{tab:maxE}
\centering
\begin{tabular}{|rl|}    \hline  
\textbf{$n$} & \textbf{$E(X | n)$} \\ \hline
10      & 1.593742 \\
50      &  1.691588 \\
100     & 1.704813 \\
500     & 1.715568 \\
1500   & 1.717376 \\
2500   & 1.717738 \\
3500   & 1.717893 \\
4500   & 1.717979 \\
10000 & 1.718145  \\
20000 & 1.718213 \\
30000 & 1.718236 \\
40000 & 1.718247  \\ \hline
\end{tabular}
\end{table}

\begin{sidewaystable*}
\captionsetup{justification=centering}
\caption{Benchmark functions.}
\label{tab:benchmark}
\centering
\small{
\begin{tabular}{lllll@{}} 
\toprule
\textbf{Name} & \textbf{Definition} & \textbf{Dim.} & \textbf{Global Minimum} 	& \textbf{Bounds}\\
\midrule
Ackley   & $f(x_1, \cdots, x_d) = -20 \exp\left(-0.2 \sqrt{\frac{1}{d} \sum_{i=1}^d x_i^2}\right) -\exp\left(\frac{1}{d} \sum_{i=1}^d \cos(2\pi x_i)\right) + 20 + e$ & 30 &\begin{tabular}{l} $f(x^{*}) = 0$ \\
  $x^{*} = (0, \cdots, 0)$ \\
\end{tabular} 
& $-10 \le x_{i} \le 10$ \\[4mm]

Rosenbrock  & 
$f(x_1, \cdots, x_d) = \sum_{i=1}^{d-1}[100 (x_{i+1} - x_i^2)^ 2 + (1 - x_i)^2]$ 
& 30 &
\begin{tabular}{l}
  $f(x^{*}) = 0$ \\
  $x^{*} = (1, \cdots, 1)$ \\
\end{tabular} 
& $-10 \le x_{i} \le 10$ \\[4mm]

Chung-Reynolds   & 
$f(x_1, \cdots, x_d) = \left(\sum_{i=1}^{d} x_{i}^{2}\right)^{2}$ 
& 30 &
\begin{tabular}{l}
  $f(x^{*}) = 0$ \\
  $x^{*} = (0, \cdots, 0)$ \\
\end{tabular} 
& $-10 \le x_{i} \le 10$ \\[4mm]

Step   & 
$f(x_1, \cdots, x_d) = \sum_{i=1}^{d} \lfloor|x_{i}|\rfloor$ 
& 30 &
\begin{tabular}{l}
  $f(x^{*}) = 0$ \\
  $x_{i}^{*} \in (-1, 1)$ \\
\end{tabular} 
& $-100 \le x_{i} \le 100$ \\[4mm]

Goldstein-Price  & 
\begin{tabular}{lll}
$f(x_1, x_2)$ & = & $\left[1 + (x_1 + x_2 + 1)^2(19 - 14x_1 +3x_1^2 - 14x_2 + 6x_1 x_2 + 3x_2^2)\right] \times$  \\
                &    & $\left[30 + (2x_1 - 3x_2)^2 (18 - 32x_1 + 12x_1^2 + 48x_2 - 36x_1 x_2 + 27x_2^2)\right]$ 
\end{tabular}
& 2 &
\begin{tabular}{l}
  $f(x^{*}) = 3$ \\
  $x^{*} = (0, -1)$ \\
\end{tabular} 
& $-2 \le x_{1}, x_{2} \le 2$ \\[4mm]
\bottomrule
\end{tabular}
}
\end{sidewaystable*}

\subsection{Empirical results}
\label{sec:empirical}
We obtain empirical estimates of the probability $p$ and the expectation $E(X | n)$ by aggregating (averaging) results from multiple, independent runs of SJaya. Table~\ref{tab:empirworst} presents the empirical average and the theoretical expectation for the functions in the benchmark test-suite in Table~\ref{tab:benchmark} (taken from \cite{chakraborty2020semi}). 
Each row in Table~\ref{tab:empirworst} corresponds to 500 runs, with each run executed for 20 generations with a specified population size. 
The theoretical expectation is obtained by plugging in the average empirical $p$ into equation~\ref{eq:theorEworst}.  
The empirical $p$ value is obtained as the average of 500 probabilities, each probability being calculated as a relative frequency from a single run, the data for a single run having been aggregated from the 20 generations comprising the run. In other words, two levels of aggregating (averaging) were implemented: aggregating over runs and aggregating over generations within a single run. While the runs are independent of one another, the generations that make up a single run are not absolutely independent, having been created on top of one another, as if in a chain or cascade. The empirical expectation of $X$ is obtained as the average (per generation per run) number of times the worst individual in the population needs to be found out.

\begin{table}
\caption{Empirical and theoretical $E(X | n)$ (rounded at the 4th decimal place)}
\label{tab:empirworst}
\centering
\begin{tabular}{|lrccc|}    \hline                                      
\textbf{Function} & \textbf{$n$} & \textbf{$p$} & \textbf{Empirical $E$} &  \textbf{Theoretical $E$}\\ 
\hline

Ackley       & 10 & 0.9230      & 1.701     &  1.4178 \\ 
                & 50 &  0.9977     & 2.0547   & 1.6855 \\
                 & 100 & 0.9985    & 2.0786 & 1.7008\\
                 & 1000 & 0.9996  & 2.1632 & 1.7158 \\ [2mm]

Rosenbrock & 10    &  0.8740  &  1.5262 &  1.3115 \\
                 & 50   & 0.9911 & 1.9779 & 1.6682 \\
                 & 100  & 0.9956 & 2.0029 & 1.6931 \\
                 & 1000 & 0.9988 & 2.0514 & 1.7137 \\ [2mm]

Chung-Reynolds & 10    & 0.9335   & 1.7408 & 1.4411 \\
                        & 50    & 0.9987   & 2.0366  & 1.6882  \\ 
                        & 100  & 0.9994   & 2.0508   & 1.7032\\
                        & 1000 & 1.0000  &  2.0984  & 1.7169 \\[2mm]

 Step & 10    & 0.9590          & 1.8392  & 1.4986 \\
        & 50    & 0.9994          & 2.0908 &  1.6900 \\
        & 100   & 0.9998         & 2.1297 &  1.7043 \\ 
        & 1000 & 1.0000          & 2.2024 &  1.7169 \\ [2mm]

 Goldstein-Price & 10    &   0.5059        & 0.6554  &  0.6381 \\
                      & 50    &      0.6286      &   0.9442 &  0.8677  \\
                      & 100    &   0.6806         &   1.1151 &  0.9705 \\
                     & 1000    &   0.7805         &  1.6763 & 1.1819  \\

\hline
\end{tabular}
\end{table}

The empirical $E$ is seen to be higher than the corresponding theoretical value in all the cases in Table~\ref{tab:empirworst}. This is explained by the fact that, in practice, the distribution described in the left side of equation~\ref{eq:transition} deviates from the uniform; what we have in practice is 
\begin{linenomath}\begin{equation}
P(W^{(t+1)} = j | W^{(t)} = k; n) > \frac{1}{n} \quad \text{for $j < k$} 
\end{equation}\end{linenomath}
where the indexing scheme is as in Figure~\ref{fig:indexed}.
This non-uniform distribution is difficult to obtain analytically. It can of course be qualitatively argued that  in a top-to-bottom processing of the population elements, the top part (comprising indices $n, n-1, \cdots, k$; $1 \le k \le n$; see Figure~\ref{fig:indexed}) gets updated before the bottom part does, and since an update never results in a worse fitness, the probability of the worst individual being found in the bottom part is higher than in the top part at any point during the course of a generation. An empirical corroboration of this can be seen in the following matrix of $P(W^{\text{(next)}} = j | W^{\text{(current)}} = k)$ values, which is obtained by averaging (in a relative-frequency sense) 5000 independent runs of SJaya on the Chung-Reynolds function (Table~\ref{tab:benchmark}) of 10 variables, where each run used 10 generations of a population of size 10:

\vspace{3mm}
\begin{center}
\begin{scriptsize}
\begin{blockarray}{cccccccccccc}
         & &  Next $\rightarrow$ &    &       &      &    &         &        &         &          &        &        \\
         & & 10  &   9  &     8  &    7  &     6  &     5  &     4  &      3  &     2  &     1 \\
\begin{block}{cc(cccccccccc)}
Current & 10	&  0.042 & 0.112 & 0.123 & 0.108 & 0.111 & 0.106 & 0.101 & 0.096 & 0.102 & 0.098 \\
$\downarrow$          &  9	&  0.089 & 0.047 & 0.118 & 0.113 & 0.117 & 0.105 & 0.105 & 0.104 & 0.103 & 0.099 \\
           &  8	&  0.098 & 0.084 & 0.042 & 0.125 & 0.116 & 0.113 & 0.108 & 0.11   & 0.104 & 0.1  \\
           &  7	&  0.099 & 0.095 & 0.096 & 0.045 & 0.113 & 0.121 & 0.118 & 0.104 & 0.105 & 0.105 \\
           &  6	&  0.102 & 0.096 & 0.092 & 0.092 & 0.048 & 0.123 & 0.111 & 0.119 & 0.11  & 0.109 \\
           &  5	&  0.1    & 0.099 & 0.097 & 0.09   & 0.092 & 0.041 & 0.128 & 0.119 & 0.121 & 0.114 \\
           &  4	&  0.11  & 0.103 & 0.105 & 0.098 & 0.096 & 0.093 & 0.04  & 0.125 & 0.112 & 0.119 \\
           &  3	 & 0.101 & 0.112 & 0.104 & 0.097 & 0.102 & 0.101 & 0.093 & 0.048 & 0.126 & 0.115 \\
           &  2	 & 0.113 & 0.108 & 0.105 & 0.102 & 0.108 & 0.095 & 0.096 & 0.098 & 0.046 & 0.129 \\
           &  1	 & 0.114 & 0.116 & 0.108 & 0.106 & 0.106 & 0.105 & 0.101 & 0.099 & 0.099 & 0.046 \\
\end{block}
\end{blockarray}
\end{scriptsize}
\end{center}
The above matrix, which, clearly, is a stochastic matrix (each row-sum is unity, ignoring floating-point errors), shows that for each row, the entries to the left of the diagonal element are smaller than those to the right of the diagonal element.

The very first or initial (before any replacement has taken place) distribution of the worstIndex, obtained from these
5000 runs and presented in Table~\ref{tab:initworstdistri}, supports the uniform distribution assumption used in equation~\ref{eq:initdistri}.

\begin{table}
\caption{Initial empirical distribution of worstIndex ($n$ = 10)}
\label{tab:initworstdistri}
\centering
\begin{tabular}{|cc|}    \hline                                      
\textbf{Index} & \textbf{Probability} \\ \hline
 10 & 0.1024 \\ 
9 & 0.0954 \\ 
8 & 0.1012 \\ 
7 & 0.0970 \\ 
6 & 0.0976 \\ 
5 &  0.0984 \\
4 &  0.0972 \\
3 &  0.1030 \\
2 &  0.1040 \\
1 &  0.1038 \\
\hline
\end{tabular}
\end{table}

\section{Updating SJaya's best-of-population index} \label{sec:bestupdate}
This section will show that the average generation-wise number of best-updates is typically small and thus does not add significantly to the computational cost of SJaya. We establish the smallness of the best-update count both empirically and  
theoretically.   

A knowledge of the expected number of updates, in a generation, of the best-of-population index is required for an analysis of SJaya. Of course, to derive this expectation, we need the underlying (discrete) probability distribution.
To compute the probability that a new individual, created in line 6 of Algorithm~\ref{algooriginal} or Algorithm~\ref{algonew},  will be better than an existing individual, we need, among other pieces of information, a knowledge of the (typically continuous) distribution of the fitness landscape. Now, the fitness distribution is impossible to know (in advance), except in trivial cases. A generic analysis, however, is possible if we are willing to make an assumption about the nature of this distribution. In the absence of any further information, we will proceed with the assumption that this distribution is normal (Gaussian). Now, the two parameters --- mean and variance --- of the normal distribution will affect the analysis quantitatively, not qualitatively. Therefore, for ease of calculations, we will use the  normal distribution with mean $\mu$ and variance $\sigma^2$.

An update of the best index is needed whenever the newly arrived individual has a fitness better than that of the current population-best. During the course of a run, the expected value of the population-best fitness at the beginning of a fresh generation 
can be modeled as the expectation of the best (either maximum or minimum, depending on the application) of $n$ i.i.d. samples drawn from a given fitness distribution, with $n$ representing the population size. (The population size is assumed not to change from generation to generation, of course.) We assume maximization without loss of generality. 

Let us use $f$ for probability density function (pdf) and $F$ for
cumulative distribution function (cdf). 
The maximum of $n$  i.i.d. samples $x_1, \cdots, x_n$ of a continuous random variable $X$ is another (continuous) random variable; call it $X_{\text{max}}$. Then the expected value of  $X_{\text{max}}$ 
is given by
\begin{linenomath}\begin{align} 
E(X_{\text{max}} \; | \; n, F_{X}) = \int_{x = -\infty}^{\infty} x\cdot f_{X_{\text{max}}}(x \; | \; n, F_{X}) \; dx   
\label{eq:Exmax}
\end{align}\end{linenomath}
where
\begin{linenomath}\begin{align} 
f_{X_{\text{max}}}(x \; | \; n, F_{X}) = n \cdot (F_{X}(x))^{n-1}   f_{X}(x)    
\end{align}\end{linenomath}
is the pdf of $X_{\text{max}}$, and 
\begin{linenomath}\begin{align}
F_{X_{\text{max}}}(x \; | \; n, F_{X}) =  P(X_{\text{max}} < x) = (F_{X}(x))^{n} 
\end{align}\end{linenomath}
is the cdf of $X_{\text{max}}$, 
such that 
\begin{linenomath}\begin{align*}
P(x < X_{\text{max}} < x+dx \; | \; n, F_{X}) = f_{X_{\text{max}}}(x \; | \; n, F_{X}) \; dx,
\end{align*}\end{linenomath}
where
\begin{linenomath}\begin{align} 
F_{X}(x) = P(X < x) = \int_{x = -\infty}^{x} f_{X}(x) \; dx   
\end{align}\end{linenomath}
is the cdf of $X$,
with $f_{X}(x)$ representing its pdf.    

To derive the expected number of updates, over a complete generation, of the population-best member, we begin by defining a discrete (binary) random variable $Y_{i, g; n, F}$ representing whether or not an update is made at iteration $i \in \{1, \cdots, n\}$ of generation $g \in \{1, \cdots, G\}$: 
\begin{linenomath}\begin{align} 
Y_{i,g; n, F} = \begin{cases}
                  1 & \quad \text{if } x > E(X_{\text{max}} \; | \; gn + i - 1, F_{X}) \\
                  0 & \quad \text{otherwise}, 
             \end{cases}   
\end{align}\end{linenomath}
with the initial generation ($g = 0$) assumed to have filled the population for the very first time. In other words, $Y_{i,g; n, F}$ is the indicator variable $1_{x > E(X_{\text{max}} \; | \; gn + i - 1, F_{X})}$. 
The expectation of $Y_{i,g; n, F}$ is given by 
\begin{linenomath}\begin{align} 
E(Y_{i,g;n,F}) =  P(x > E(X_{\text{max}} \; | \; gn + i - 1, F_{X}))   
\end{align}\end{linenomath}
If $Y_{g;n,F}$ denotes a random variable representing the total number of updates in a given generation $g$, we have 
\begin{linenomath}\begin{align} 
Y_{g;n,F} =  \sum_{i=1}^{n} Y_{i,g;n,F},   
\end{align}\end{linenomath}
and the expectation of $Y_{g;n,F}$ is then obtained as
\begin{linenomath}\begin{align} 
E(Y_{g;n,F}) = & E\left(\sum_{i=1}^{n}Y_{i,g;n,F}\right) \\
              = & \sum_{i=1}^{n} E(Y_{i,g;n,F})  \\
              =  & \sum_{i=1}^{n} P(x > E(X_{\text{max}} \; | \; gn + i - 1, F_{X})) \label{eq:EY}
\end{align}\end{linenomath}
where linearity of expectation has been used (the linearity is applicable regardless of whether or not the $Y_{i,g;n,F}$'s are independent). 

From the definition of $X_{\text{max}}$ it follows that for $n_2 > n_1$, 
\begin{linenomath}\begin{align*}
E(X_{\text{max}} \; | \; n_{2}, F_{X}) >  E(X_{\text{max}} \; | \; n_{1}, F_{X}),
\end{align*}\end{linenomath}
which implies
\begin{linenomath}\begin{align*}
P(x > E(X_{\text{max}} \; | \; n_{2}, F_{X})) <  P(x > E(X_{\text{max}} \; | \; n_{1}, F_{X})).
\end{align*}\end{linenomath}

Thus we have
\begin{linenomath}\begin{align}
E(X_{\text{max}} \; | \; g_{2}n + i - 1, F_{X}) >  E(X_{\text{max}} \; | \; g_{1}n + i - 1, F_{X}) \text{ for } g_2 > g_1 \ge 1
\label{eq:first}
\end{align}\end{linenomath}
or equivalently, 
\begin{linenomath}\begin{align}
 E(Y_{i,g_{2};n,F}) < E(Y_{i,g_{1};n,F})        \text{ for } g_2 > g_1 \ge 1. 
\end{align}\end{linenomath}
Again 
\begin{linenomath}\begin{align} 
E(X_{\text{max}} \; | \; gn + i_{2} - 1, F_{X}) >  E(X_{\text{max}} \; | \; gn + i_{1} - 1, F_{X}) \text{ for } i_2 > i_1,
\end{align}\end{linenomath}
or equivalently, 
\begin{linenomath}\begin{align}
 E(Y_{i_{2},g;n,F}) < E(Y_{i_{1},g;n,F}) \text{ for } i_2 > i_1. \label{eq:second}
\end{align}\end{linenomath}
Therefore
\begin{linenomath}\begin{align} 
E(X_{\text{max}} \; | \; (g+1)n, F_{X}) >  E(X_{\text{max}} \; | \; gn + n - 1, F_{X})  \label{eq:third}
\end{align}\end{linenomath}
which shows that the $E(Y_{i,g;n,F})$ value corresponding to the last iteration of any generation is strictly greater than that corresponding to the first iteration of the immediately following generation.
Inequalities \ref{eq:first}-\ref{eq:third} lead to 
\begin{linenomath}\begin{align*}
E(Y_{g_{2};n,F}) < E(Y_{g_{1};n,F}) \quad \text{for } g_2 > g_1 \ge 1. 
\end{align*}\end{linenomath}
Note that the above inequality holds for any $F_{X}$ (or equivalently, for any $f_{X}$) and for any $n$. Thus we have proved the following theorem:

\vspace{3mm}
\noindent
{\bf Theorem 2}: {\em For any problem, in any run, the expected generation-wise best-update count decreases monotonically with generations, regardless of the population size.}

\vspace{3mm}
Let 
\begin{linenomath}\begin{displaymath}
\bar{Y} = \frac{1}{G}\sum_{g=1}^{G} E(Y_{g;n,F})
\end{displaymath}\end{linenomath}
 stand for the average (over all the generations in a run) of the expected  generation-wise best-update counts. Then
Theorem 2 implies that $\bar{Y}$
 can be made arbitrarily small by making the total number of generations $G$ arbitrarily large, a fact that allows us to argue that the number of best-updates per generation (or per run) should not be a concern, so far as computational costs are considered. While that argument is theoretically sound ($\bar{Y}$ does indeed $\to$ 0 as $G \to \infty$) and empirical results (Section~\ref{sec:empirbestupd}) show that the $E(Y_{1;n,F})$ is small even for large n and that the update count drops fast with generations, the caveat is that because the true density $f_{X}$ (or equivalently, the true cdf $F_{X}$) always remains unknown and because for an arbitrary $f_{X}$, it is difficult, if not impossible, to obtain a tight upper bound on $E(Y_{1;n,F})$, a proof that the average generation-wise best-update count is guaranteed, {\em regardless of the problem}, to drop to a specified (small) value after the consumption of a specified (modest) number of generations remains elusive. Below we consider four particular distributions for which we establish upper bounds on $E(Y_{1;n,F})$; these four cases are potential candidates for approximations to the true (unknown) distributions.

\subsection{Special cases}
\subsubsection{The uniform random distribution}  
For the $\text{Unif}(a,b)$ distribution, the density is given by 
\begin{linenomath}\begin{align}
f_{X}(x) = \frac{1}{b-a}; \quad  b > a, \quad x \in [a, b]
\end{align}\end{linenomath}
and the corresponding cdf is
\begin{linenomath}\begin{align}
F_{X}(x) = \frac{x-a}{b-a},
\end{align}\end{linenomath} 
which, when used in equation~\ref{eq:Exmax}, gives
\begin{linenomath}\begin{align}
E(X_{\text{max}} \; | \; n, \text{Unif}_{X}) = \frac{a+bn}{n+1},
\end{align}\end{linenomath} 
from which we get 
\begin{linenomath}\begin{align}
P(x > E(X_{\text{max}} \; | \; n , \text{Unif}_{X})) = \frac{1}{n+1}. 
\end{align}\end{linenomath} 
This expression allows us to obtain $E(Y_{1;n, \text{Unif}})$ from equation~\ref{eq:EY} as
\begin{linenomath}\begin{align}
E(Y_{1;n, \text{Unif}}) = & \sum_{j = 0}^{n-1} \frac{1}{n+j+1} \\
                   = & H_{2n} - H_{n}
\end{align}\end{linenomath} 
where $H_{n}$ is the $n$-th harmonic number.
It is not difficult to prove from either of the above two equations that $E(Y_{1;n, \text{Unif}})$ is monotone increasing in $n$.
Luckily, an upper bound on $E(Y_{1;n,\text{Unif}})$ can be obtained using the fact that 
\begin{linenomath}\begin{align}
\lim_{n \to \infty} \left(  H_{2n} - H_{n} \right) = \ln 2. 
\end{align}\end{linenomath}
Thus, for any $n$, no matter how large, and any $a$ and $b$
\begin{linenomath}\begin{align}
E(Y_{1;n,\text{Unif}}) \le \ln 2. 
\end{align}\end{linenomath}
The minimum value of the expectation is 1/2 and corresponds to $n=1$ (recall that the mean of the $\text{Unif}(a.b)$ distribution is $(a+b)/2$ and that the area under the pdf box to the right of $(a+b)/2$ is 1/2).

Theoretical expectations of $Y_1$ values corresponding to different population sizes are presented in Table~\ref{tab:Ebounds} where the corresponding values for three other distributions are also shown.

\subsubsection{The exponential distribution}
For the exponential distribution, the pdf and cdf are given by 
\begin{linenomath}\begin{align}
f_{X}(x) = \lambda e^{-\lambda x}; \quad  \lambda > 0,  \quad x \in [0, +\infty)
\end{align}\end{linenomath}
and
\begin{linenomath}\begin{align}
F_{X}(x) = 1 - e^{-\lambda x} 
\end{align}\end{linenomath}
which lead to
\begin{linenomath}\begin{align}
E(X_{\text{max}} \; | \; n, \text{Exp}_{X}) = & \int_{x = 0}^{\infty}  x \cdot n \cdot \lambda e^{-\lambda x} (1 - e^{-\lambda x})^{n-1}\; dx \\
                                                              = & \frac{1}{\lambda} \sum_{k=1}^{n} \frac{1}{k} \\
                                                              = & \frac{1}{\lambda} H_{n}
\end{align} \end{linenomath}
We then have
\begin{linenomath}\begin{align}
P(x > E(X_{\text{max}} \; | \; n , \text{Exp}_{X})) = &  e^{-\lambda \times \frac{1}{\lambda} H_{n}}  =  e^{- H_{n}}               
\end{align}\end{linenomath}
from which we obtain (by equation~\ref{eq:EY})
\begin{linenomath}\begin{align}
E(Y_{1;n, \text{Exp}}) = & \sum_{j = 0}^{n-1}  e^{- H_{n+j}}          
\label{eq:expY}         
\end{align} \end{linenomath}
Now, 
$E(Y_{1;n, \text{Exp}})$ is monotone non-decreasing in $n$. Thus the smallest value of this expectation occurs at $n=1$, and that value, from equation~\ref{eq:expY}, is $1/e$ or 0.3679, a value that is corroborated by the fact that the mean of the exponential distribution is $1/\lambda$ and that the area under the pdf to the right of the point $1/ \lambda$ is $e^{- \lambda \times (1/\lambda)}$ or $1/e$.  

The case corresponding to an arbitrarily large $n$ can be studied by using the fact that 
\begin{linenomath}\begin{align}
\lim_{n \to \infty} (H_n - \ln n) = \gamma,
\label{Euler-Masch}
\end{align} \end{linenomath}
where $\gamma \approx 0.5772$ is the Euler-Mascheroni constant.
From equation~\ref{eq:expY}, we have
\begin{linenomath}\begin{align}
\lim_{n \to \infty} E(Y_{1;n, \text{Exp}}) =  & \lim_{n \to \infty} \sum_{j = 0}^{n-1}  e^{-\ln(n+j) - \gamma}           \\        
                                                          =   & \frac{1}{e^{\gamma}} \lim_{n \to \infty}  \sum_{j = 0}^{n-1}  \frac{1}{n+j}   \\
                                                          =  &   \frac{1}{e^{\gamma}} \ln 2             
\end{align}\end{linenomath} 
Thus, for any $n$ and any $\lambda$, 
\begin{linenomath}\begin{align}
E(Y_{1;n,\text{Exp}}) \le \frac{\ln 2}{e^{\gamma}} = 0.3892. 
\end{align}\end{linenomath}
Table~\ref{tab:Ebounds} shows how the theoretical $E(Y_{1;n,\text{Exp}})$ varies with $n$, reaching the limit as $n$ approaches infinity. 

\subsubsection{The normal distribution}
The normal distribution $N(\mu, \sigma)$, with mean $\mu$ and standard deviation $\sigma$, unfortunately, admits of no closed-form expression for $E(X_{\text{max}} \; | \; n, F_{X})$ when 
\begin{linenomath}\begin{align}
f_{X}(x) =  \frac{1}{\sigma \sqrt{2 \pi}} e^{-\frac{1}{2} \left(\frac{x-\mu}{\sigma}\right)^{2}}; \quad \sigma > 0, \quad x \in (-\infty, +\infty)
\end{align} \end{linenomath}
and the integration in equation~\ref{eq:Exmax} must be evaluated numerically.
We find $E(Y_{1;n, \text{Norm}})$ numerically, from equation~\ref{eq:EY} (using numerical routines from Python's {\em scipy} \cite{2020SciPy-NMeth, 2020NumPy-Array} and also from Mathematica  \cite{Mathematica}). The numerically obtained $Y_1$ values corresponding to different population sizes are presented in Table~\ref{tab:Ebounds}. Note that $Y_{1}$ is monotone non-increasing with $n$ 
for the Gaussian. Thus the maximum possible value of $E(Y_{1;n,\text{Norm}})$ is obtained at the smallest possible value of $n$, namely 1, and the corresponding $E(X_{\text{max}})$ is clearly the mean, $\mu$, of the distribution, which, because of symmetry (the mean equals the median), causes 
$P(x > E(X_{\text{max}} \; | \; 1, \text{Norm}_{X}))$ to be 0.5, leading to $E(Y_{1;1,\text{Norm}})  = 0.5$.
Thus, for any $n$, no matter how large, and any $\mu$ and $\sigma$, $E(Y_{1;n,\text{Norm}})$ is upper-bounded by
\begin{linenomath}\begin{align}
E(Y_{1;n,\text{Norm}}) \le 0.5. 
\end{align}\end{linenomath}

The limit of the expectation, as $n \to \infty$, cannot be obtained analytically.

\subsubsection{The logistic distribution}
The logistic distribution offers some similarity (e.g., unimodality, symmetry) to the normal. That, coupled with the fact that it is amenable to analytical treatment, affords an alternative to the normal distribution for modeling purposes. For simplicity, let us consider location and scale parameters of 0 and 1, respectively. This does not cause any loss of generality, because any logistic variable $X$ with location $a$ and scale $s > 0$ can be transformed to another logistic variable $Z$:
\begin{linenomath}\begin{align}
Z = \frac{X - a}{s}.
\end{align}  \end{linenomath}
The cdf and pdf of the logistic distribution are given by
\begin{linenomath}\begin{align}
F_{X}(x) = \frac{1}{1 + e^{- x}}; \quad x \in (-\infty, +\infty)  
\end{align}\end{linenomath}
and
\begin{linenomath}\begin{align}
f_{X}(x) = F_{X}(x) (1 - F_{X}(x))  = \frac{e^{- x}}{(1 + e^{- x})^{2}}. 
\end{align}\end{linenomath}
Thus
\begin{linenomath}\begin{align}
E(X_{\text{max}} \; | \; n, \text{Logistic}_{X}) = &  \int_{x = -\infty}^{\infty}  \frac{n \cdot x  \cdot  e^{- x}}{ (1 + e^{- x})^{n+1}}\; dx \\
                                                              = \; & H_{n-1},  
\end{align} \end{linenomath}
with $H_{0}$ taken to be zero (recall that $n \ge 1$). Then
\begin{linenomath}\begin{align}
P(x > E(X_{\text{max}} \; | \; n , \text{Logistic}_{X})) = &  \frac{e^{- H_{n-1}}}{1+e^{- H_{n-1}}}  
\end{align}\end{linenomath}
and, by equation~\ref{eq:EY}, 
\begin{linenomath}\begin{align}
E(Y_{1;n, \text{Logistic}}) = \sum_{j = 0}^{n-1}  \frac{e^{- H_{n+j-1}}}{1+e^{- H_{n+j-1}}}. 
\label{eq:logisticY}                
\end{align} \end{linenomath}

$E(Y_{1;n, \text{Logistic}})$ is monotone non-increasing in $n$. 
Thus the largest value of the expectation occurs at $n=1$, and that expectation is obtained from equation~\ref{eq:logisticY} as 0.5, a value that is corroborated by the symmetric nature of the distribution. Now, using the large-$n$ approximation to $H_n$, namely
$\lim_{n \to \infty} H_n = \ln n +  \gamma$ (recall equation~\ref{Euler-Masch}),
we have
\begin{linenomath}\begin{align*}
\lim_{n \to \infty} E(Y_{1;n, \text{Logistic}}) =  & \lim_{n \to \infty} \sum_{j = 0}^{n-1}  \frac{e^{-\ln(n+j-1) - \gamma}}
                                                                                                                               {1 + e^{-\ln(n+j-1) - \gamma}}           \\        
                                                          =   & \frac{1}{e^{\gamma}} \lim_{n \to \infty} \sum_{j = 0}^{n-1} \frac{1}{n+j-1 + e^{-\gamma}}   \\
                                                          =  &   \frac{1}{e^{\gamma}} \ln 2   \\
                                                          =  & 0.3892.      
\end{align*}\end{linenomath}

\begin{table}
\caption{Growth (or decay) of $E(Y_{1;n,F})$ with $n$}
\label{tab:Ebounds}
\centering
\begin{tabular}{|rcccc|}    \hline                                      
\multirow{2}{*}{$n$}    &       \multicolumn{4}{c|}{$E(Y_{1;n,F})$} \\
\cline{2-5}
         &        $F = \text{Exp}(\lambda)$ &    $F = \text{Logistic(0,1)}$ &  $F = N(\mu, \sigma)$    &   $F = \text{Unif}(a,b)$ \\ \hline
1       &        0.3679 &  0.5                     & 0.5             &  0.5 \\ 
10     &        0.3889           &  0.4016                & 0.4451       & 0.6688 \\ 
50     &        0.3892          &  0.3916                & 0.4261       &  0.6882\\ 
100   &        0.3892           &  0.3904                & 0.4212        & 0.6907\\ 
500   &        0.3892           & 0.3894                 & 0.4136       & 0.6926 \\ 
5000 &        0.3892           &   0.3892               & 0.4074       & 0.6931\\ 
10000 &       0.3892           &   0.3892              & 0.4061       &  0.6931  \\
$\infty$  &  0.3892   &  0.3892               &  ---              &  0.6931\\ 
\hline
\end{tabular}
\end{table}

\subsection{Empirical results} \label{sec:empirbestupd}
Empirical values of the average counts of best-updates are obtained by aggregating independent SJaya runs for each of the test functions. For a population size of $n$, the empirical expectation (average) at a given generation $g$ is produced from an ensemble of $r$ runs as:
\begin{linenomath}\begin{align}
 E_{\text{empir}}(Y_{g;n}) = \frac{1}{r} \sum_{k=1}^{r} N_{k}(g)
\end{align}\end{linenomath}
where $N_{k}(g)$ is the number (an integral count $\ge 0$) of updates of the best-index at generation $g$ in run $k$. The corresponding average (over all generations) is obtained as
\begin{linenomath}\begin{align}
 E_{\text{empir}}(Y_{n}) = \frac{1}{G} \sum_{g=1}^{G} E_{\text{empir}}(Y_{g;n}).
\end{align}\end{linenomath}
In the above two equations, $g \ge 1$ ($g = 0$ represents the initial   
population).
Table~\ref{tab:empirEYbest} shows, for different population sizes, $E_{\text{empir}}(Y_{n})$ values as well as how 
 $E_{\text{empir}}(Y_{g;n})$ changes with generations ($r = 500$ and $G = 20$ in this table).  The runs used in this table are the same as the ones used in Table~\ref{tab:empirworst}.

\begin{table}
\caption{$E_{\text{empir}}(Y_{g;n})$ for $g = 1, 10, 20$ (number of runs = 500)}
\label{tab:empirEYbest}
\centering
\begin{tabular}{|lrcccc|}    \hline                                      
\multirow{2}{*}{Function}           &   \multirow{2}{*}{$n$}    & \multicolumn{3}{c}{$E_{\text{empir}}(Y_{g;n})$}  & \multirow{2}{*}{$E_{\text{empir}}(Y_{n})$} \\
\cline{3-5} 
                       &            &   Gen 1    & Gen 10    & Gen 20   &        \\ \hline

Ackley            &  10     & 0.916 & 0.484  & 0.452 & 0.5276\\ 
                      & 50      & 0.488 & 0.278 & 0.184 & 0.2882 \\
                      & 100      & 0.352 & 0.216  & 0.182  & 0.2261\\
                      &  1000     & 0.130 & 0.156 & 0.122 & 0.1495\\[2mm]

Rosenbrock      &  10      & 0.650   & 0.396  & 0.404 & 0.445 \\
                      &   50     & 0.254    & 0.230  & 0.186    & 0.2275 \\
                      &   100     & 0.206   & 0.168   & 0.112   & 0.1763 \\
                      &  1000    & 0.052    & 0.104  &  0.090   & 0.0868 \\  [2mm]

Chung-Reynolds & 10      & 0.854    &  0.46  & 0.49                                      & 0.5353 \\   
                        & 50      & 0.418 & 0.232 & 0.192 & 0.2852 \\
                        & 100    & 0.300   &  0.174          & 0.124    & 0.2148 \\ 
                        & 1000  & 0.084 &  0.146    & 0.148  & 0.1386 \\  [2mm]

Step                 & 10 & 0.904         &  0.506          &    0.516         & 0.5858 \\
                       & 50 & 0.468 & 0.276 & 0.236 & 0.3116\\
                       & 100  & 0.400 &   0.222 &  0.154   &  0.2614\\
                       & 1000  & 0.148 & 0.204 & 0.106  & 0.1836\\[2mm]

Goldstein-Price & 10  &  0.600 & 0.170 & 0.076 & 0.208 \\
                      & 50  &  0.610 & 0.138 & 0.058 & 0.196 \\ 
                      & 100  & 0.620 & 0.142 & 0.009 & 0.1942 \\
                      & 1000  & 0.588 & 0.150 & 0.074 & 0.1802 \\
\hline
\end{tabular}
\end{table}

\section{Computational costs}
While the Jaya algorithm finds the best-of-population member and the worst-of-population member exactly once per generation, SJaya does this on a continuous, as-needed basis. The logic for finding the best (or worst) of a given number of elements can be implemented as a simple sequential scan of the elements, consisting of two basic operations for each element: a comparison followed, conditionally, by an assignment. We now find the costs of these two types of operations.

\subsection{Comparison and assignment operations for best-index update in SJaya}
\label{sec:compassignsjaya}
The total number of  comparison operations needed for updating the best index in an entire generation of SJaya (call this number $\#\text{comp}$) is equal to the number of times line 9 in Algorithm~\ref {algonew} is executed (i.e., the condition in line 9 is tested) per generation (this number is the same as the number of times the condition in line 7 evaluates to TRUE in a generation). We need to find the expected value of $\#\text{comp}$.

We can model the new individual being at least as good as the current individual (in line 7) by the event $X_2 \ge X_1$, where $X_1$ and $X_2$ are two independent random samples drawn from the same (unknown) distribution. (As mentioned earlier, this distribution is never truly known, and we have to make do with estimates and approximations.)

Defining a random variable 
\begin{linenomath}\begin{align}
Z = 1_{X_2 \ge X_1},
\end{align}\end{linenomath}
we have
\begin{linenomath}\begin{align}
Z = \begin{cases}
       1 & \text{ with probability } P(X_2 \ge X_1) \\
       0 & \text{ with probability } 1 - P(X_2 \ge X_1).
       \end{cases}
\end{align}\end{linenomath}
Thus 
\begin{linenomath}\begin{align}
E(Z) = P(X_2 \ge X_1),
\end{align}\end{linenomath}
and the expectation of the total number of comparison operations in a generation of SJaya is given by
\begin{linenomath}\begin{align}
E(\#\text{comp}) = &  E\left(\sum_{i=1}^{n} Z_i\right) \\
                = &  \sum_{i=1}^{n} E(Z_i)  \quad \text{(by linearity)}\\
                = & n \: E(Z)
\end{align}\end{linenomath}
where the last step follows from the fact that the events at the $n$ slots of the population are governed by the same underlying distribution.
Thus 
\begin{linenomath}\begin{align}
E(\#\text{comp})  = &  n \: P(X_1 \le X_2), 
\end{align}\end{linenomath}
which, by the law of total probability, gives
\begin{linenomath}\begin{align}
E(\#\text{comp})  = & n \: \int_{x = -\infty}^{\infty} P(X_1 \le X_2 | X_2 = x) \: f_{X_2}(x) \: dx \\
                          = & n \: \int_{x = -\infty}^{\infty} P(X_1 \le x) \: f_{X_2}(x) \: dx \\
                          = & n \: \int_{x = -\infty}^{\infty} F_{X_1}(x) \: f_{X_2}(x) \: dx \\
                          = & n \: \int_{x = -\infty}^{\infty} F_{X}(x) \: f_{X}(x) \: dx
\end{align}\end{linenomath}

If the distribution of $X$ is known, we can analytically obtain $E(\#\text{comp})$. For instance, this expectation is $n/2$ for exponential, uniform and logistic distributions but cannot be obtained in an explicit closed form for the normal distribution.

Next, the total number of  assignment operations (call it $\#\text{assign}$) needed for updating the best index in an entire generation of SJaya is equal to the number of times line 10 is executed per generation in Algorithm~\ref {algonew}. The expectation of this count, $E(\#\text{assign})$, was already derived in Section~\ref{sec:bestupdate}; $E(\#\text{assign})$ can be taken to be either the generation-specific $E(Y_{g;n,F})$ or the average $\bar{Y}$. This expectation is obviously a function of the population size $n$, and Section~\ref{sec:bestupdate} obtained the maximum value of this expectation (corresponding to either $n = 1$ or $n \to \infty$, depending on the nature of the distribution) for specific distributions, as follows:
\begin{linenomath}\begin{align}
\text{Max. of } E(\#\text{assign}) = \begin{cases}
       (\ln 2) / e^{\gamma}  & \text{ for exponential distribution} \\
        \ln 2 & \text{ for uniform distribution} \\
       0.5 & \text{ for normal distribution} \\
       0.5 & \text{ for logistic distribution}.
       \end{cases}
\end{align} \end{linenomath}
As mentioned earlier, it is difficult to obtain a closed-form analytical expression of this expectation for arbitrary distributions; however, by Theorem 2, this expectation, when averaged over a number of generations, goes down as the number of generations increases, regardless of the underlying distribution.

\subsection{Comparison and assignment operations for finding the best/worst of $n$ elements}
The na\"{\i}ve approach to sequentially scanning an array for finding the best (or worst) element entails exactly $n$ (or $n-1$, depending on the implementation) comparisons:
\begin{linenomath}\begin{align}
\#\text{comp\_naive} =  n.
\end{align}\end{linenomath}
The number of assignments, however, is not deterministic. 
Assuming the array index runs from 1 to $n$, the number of assignments can go from a minimum of 1 to a maximum of $k$ (or from 1 to $n-k+1$, depending on the implementation), inclusive, when the best (or worst) element is located at index $k$. The average-case analysis can be performed by noting that when the numbers are uniformly distributed in the array, the $j$-th element is greater (smaller) than the preceding $j-1$ elements with probability $1/j$, independently for all $j = 1, \cdots,n$.  
Thus
\begin{linenomath}\begin{align}
E(\#\text{assign\_naive}) = &  \sum_{j=1}^{n} \frac{1}{j}  \\
                                            = & H_n
\end{align}\end{linenomath}

\subsection{Complexity of SJaya}
Given the analyses of the preceding sections, it is now straightforward to obtain the complexity of SJaya.
The cost of the initialization step (line 1 in Algorithm~\ref{algonew})  is $n \times \phi(d)$, where $\phi(d)$ is the cost of evaluating the fitness (objective function) of a given problem of $d$ dimensions. The cost of finding the best/worst member in the entire population (line 2) is $E(\#\text{comp\_naive}) \times C_c + E(\#\text{assign\_naive})  \times C_a$ or $n \times C_c + H_n \times C_a$, where $C_c$ and $C_a$ are the cost of a single comparison and a single assignment, respectively. Setting the random parameters for the solution vector (line 4) has a cost of $C_{p} \times d$, where $C_p$ is a constant. Creating a single new individual (line 6) incurs a cost of  $C_{\text{op}} \times d + \phi(d)$, where $C_{\text{op}}$ represents the cost per dimension of applying the algebraic operations involved in the creation of a new individual. Each check for the superiority of the new individual (line 7) costs $C_c$, and there are $n$ such checks in a generation. The replacement at line 8 takes place  
$E(\#\text{comp})$ times in a whole generation (recall that the condition at line 7 is true these many times on average in a generation). Again, the condition in line 9 is tested 
$E(\#\text{comp})$ times in a whole generation. And, as already shown in Sec.~\ref{sec:compassignsjaya}, the update in line 10 occurs $E(\#\text{assign})$ times per generation. The condition in line 12 is tested $E(\#\text{comp})$ times in a generation, and finding the worst individual in line 13 is needed a maximum of 1.7 times per generation. 
 The total cost of a single run is thus
\begin{linenomath}\begin{align*}
& n \phi(d)  +  2(n C_c + H_n C_a) +  G \Big[C_{p}d  + n(C_{\text{op}}d + \phi(d)) + n C_c  + E(\#\text{comp}) \; C_a \\&                   
                  + E(\#\text{comp}) \; C_c + E(\#\text{assign}) \; C_a
                                    + E(\#\text{comp}) \; C_c + 1.7 (n C_c + H_n C_a)
                 \Big]. 
\end{align*}\end{linenomath}

\subsection{Complexity of Jaya}
The complexity of Jaya can now be derived easily. Most of the calculations carry over from those of SJaya. Noting that the replacement of the existing individual with the new one (in line 8 of Algorithm~\ref{algooriginal}) takes place $E(\#\text{comp})$
times in an entire generation, the complexity is given by
\begin{linenomath}\begin{align*}
& n \phi(d)  +   G \Big[ 2(n C_c + H_n C_a) +  C_{p}d  + n(C_{\text{op}}d + \phi(d)) + n C_c  + E(\#\text{comp}) \; C_a \Big]. 
\end{align*}\end{linenomath}

\subsection{Cost difference between SJaya and Jaya}
Using the upper bound of $\#\text{assign}$, namely $\#\text{comp}$, and assuming $E(\#\text{comp}) = n/2$, we obtain the following estimate of an upper bound of the additional cost incurred by SJaya over Jaya per generation:
\begin{linenomath}\begin{align}
\text{additional cost} \le & \; \frac{n}{2}(2C_a + C_c) - 0.3 (n C_c + H_n C_a) \\
                              \approx & \; (n - 0.3\ln n - 0.17316) C_a + 0.2 n C_c \quad \text{for large $n$}.
\end{align}\end{linenomath}
This additional cost is not significant compared to the total cost of evaluating the fitnesses of the $n$ population members in a generation. If, in light of the analysis in Sec.~\ref{sec:bestupdate}, a more realistic value of $E(\#\text{assign}) \ll n$ is assumed, the additional cost becomes even lower.   

\section{Conclusions}
A theoretical analysis of two stochastic heuristics --- Jaya and its recent improvement SJaya --- has been presented in this paper. 
A remarkable fact revealed by the analysis 
is that the maximum expected number of worst-index updates per generation for SJaya is only about 1.7 for almost any population size of practical interest. 
Furthermore, regardless of the population size, the expectation of the number of best-index updates per generation decreases monotonically with generations.
We derived exact upper bounds of the expected number of best-index updates when the underlying distribution is exponential, logistic, normal or uniform.  Asymptotics of expected best-update counts were obtained for exponential, logistic and uniform distributions. Limitations of the analytical approach and the need to resort on occasion to numerical techniques have been pointed out. The model allowed us to obtain computational complexities of the algorithms, which showed that the performance improvement afforded by SJaya over Jaya incurs only a modest additional cost. Empirical results on benchmark test problems were obtained and found to corroborate the theoretical findings.  To our knowledge, this is the first theoretical analysis of this powerful and popular family of heuristics. The insights provided by our models should help design new, improved  population-based search/optimization heuristics in machine learning / artificial intelligence.  
The analytical approach developed here has the potential to be extended to the analysis of other types of evolutionary algorithms. 

\section*{Acknowledgements}
Thanks to Adrian Clingher for his help with the derivation of the second identity used in the proof of Theorem 1. Mike Toohey provided help over a weekend, scheduling servers to accommodate my CPU-intensive jobs.

\bibliographystyle{ieeetr}
\bibliography{udaybib2}

\begin{thebibliography}{10}

\bibitem{rao2016jaya}
R.~V. Rao, ``Jaya: A simple and new optimization algorithm for solving
  constrained and unconstrained optimization problems,'' {\em International
  Journal of Industrial Engineering Computations}, vol.~7, no.~1, pp.~19--34,
  2016.

\bibitem{rao2019jaya}
R.~V. Rao, {\em Jaya: An Advanced Optimization Algorithm and its Engineering
  Applications}.
\newblock Springer, 2019.

\bibitem{venkata2020multi}
R.~V. Rao, A.~Saroj, J.~Taler, and P.~Oclon, ``Multi-objective design
  optimization of shell-and-tube heat exchanger using multi-objective
  {SAMP}-{J}aya algorithm,'' in {\em Advanced Engineering Optimization Through
  Intelligent Techniques}, pp.~831--838, Springer, 2020.

\bibitem{tripathy2021jaya}
S.~Tripathy, M.~K. Debnath, and S.~K. Kar, ``Jaya algorithm tuned {FO-PID}
  controller with first order filter for optimum frequency control,'' in {\em
  2021 1st Odisha International Conference on Electrical Power Engineering,
  Communication and Computing Technology (ODICON)}, pp.~1--6, IEEE, 2021.

\bibitem{yadav2022effective}
R.~Yadav and R.~Panwar, ``Effective medium approximation fused optimization
  strategy derived new kind of honeycomb microwave absorbing structure,'' {\em
  IEEE Transactions on Magnetics}, 2022.

\bibitem{agarwal2018new}
N.~Agarwal, M.~Pradhan, and N.~Shrivastava, ``A new multi-response {J}aya
  algorithm for optimisation of {EDM} process parameters,'' {\em Materials
  Today: Proceedings}, vol.~5, no.~11, pp.~23759--23768, 2018.

\bibitem{gupta2021efficient}
S.~Gupta, N.~Kumar, and L.~Srivastava, ``An efficient {J}aya algorithm with
  powell’s pattern search for optimal power flow incorporating distributed
  generation,'' {\em Energy Sources, Part B: Economics, Planning, and Policy},
  vol.~16, no.~8, pp.~759--786, 2021.

\bibitem{chakraborty2019proton}
U.~K. Chakraborty, ``Proton exchange membrane fuel cell stack design
  optimization using an improved {J}aya algorithm,'' {\em Energies}, vol.~12,
  no.~16, p.~3176, 2019.

\bibitem{satapathy2018jaya}
S.~C. Satapathy and V.~Rajinikanth, ``Jaya algorithm guided procedure to
  segment tumor from brain {MRI},'' {\em Journal of Optimization}, 2018.

\bibitem{mohapatrajaya}
P.~Mohapatra, R.~Mishra, and T.~K. Patra, ``A {J}aya algorithm trained neural
  network for stock market prediction,'' {\em International Journal of
  Innovative Technology and Exploring Engineering}, vol.~7, pp.~9--13, 2018.

\bibitem{zitar2021intensive}
R.~A. Zitar, M.~A. Al-Betar, M.~A. Awadallah, I.~A. Doush, and K.~Assaleh, ``An
  intensive and comprehensive overview of {JAYA} algorithm, its versions and
  applications,'' {\em Archives of Computational Methods in Engineering},
  vol.~28, pp.~1--30, 2021.

\bibitem{chakraborty2020semi}
U.~K. Chakraborty, ``Semi-steady-state {J}aya algorithm for optimization,''
  {\em Applied Sciences}, vol.~10, no.~15, p.~5388, 2020.

\bibitem{2020SciPy-NMeth}
P.~Virtanen, R.~Gommers, T.~E. Oliphant, M.~Haberland, T.~Reddy, D.~Cournapeau,
  E.~Burovski, P.~Peterson, W.~Weckesser, J.~Bright, S.~J. {van der Walt},
  M.~Brett, J.~Wilson, K.~J. Millman, N.~Mayorov, A.~R.~J. Nelson, E.~Jones,
  R.~Kern, E.~Larson, C.~J. Carey, {\.I}.~Polat, Y.~Feng, E.~W. Moore,
  J.~{VanderPlas}, D.~Laxalde, J.~Perktold, R.~Cimrman, I.~Henriksen, E.~A.
  Quintero, C.~R. Harris, A.~M. Archibald, A.~H. Ribeiro, F.~Pedregosa, P.~{van
  Mulbregt}, and {SciPy 1.0 Contributors}, ``{{SciPy} 1.0: Fundamental
  Algorithms for Scientific Computing in Python},'' {\em Nature Methods},
  vol.~17, pp.~261--272, 2020.

\bibitem{2020NumPy-Array}
C.~R. Harris, K.~J. Millman, S.~J. van~der Walt, R.~Gommers, P.~Virtanen,
  D.~Cournapeau, E.~Wieser, J.~Taylor, S.~Berg, N.~J. Smith, R.~Kern, M.~Picus,
  S.~Hoyer, M.~H. van Kerkwijk, M.~Brett, A.~Haldane, J.~Fernández~del Río,
  M.~Wiebe, P.~Peterson, P.~Gérard-Marchant, K.~Sheppard, T.~Reddy,
  W.~Weckesser, H.~Abbasi, C.~Gohlke, and T.~E. Oliphant, ``Array programming
  with {NumPy},'' {\em Nature}, vol.~585, p.~357–362, 2020.

\bibitem{Mathematica}
W.~R. Inc., ``Mathematica, {V}ersion 12.3.''
\newblock Champaign, IL, 2021.

\end{thebibliography}

\end{document}